\documentclass{article}
\usepackage{spconf}
\usepackage{graphicx}
\usepackage{booktabs}
\usepackage{amsmath}
\usepackage{hyperref}
\usepackage{multirow}
\usepackage{fixltx2e}

\title{FCHD: Fast and accurate head detection in crowded scenes }
\twoauthors
 {Aditya Vora}
	{Senior Data Scientist,\\
	Honeywell Technology Solutions\\
	Bengaluru, Karnataka, India - 560048}
 {Vinay Chilaka}
	{Indian Institute of Technology, Madras\\
	Computer Science Department\\
	Chennai, Tamil Nadu, India - 600036}

\begin{document}
%\ninept
%
\maketitle
\begin{abstract}
\label{sec:abstract}
In this paper, we propose FCHD-Fully Convolutional Head Detector, an end-to-end trainable head detection model. Our proposed architecture is a single fully convolutional network which is responsible for both bounding box prediction and classification. This makes our model lightweight with low inference time and memory requirements. Along with run-time, our model has better overall average precision (AP) which is achieved by selection of anchor sizes based on the effective receptive field of the network. This can be concluded from our experiments on several head detection datasets with varying head counts. We achieve an AP of $0.70$ on a challenging head detection dataset which is comparable to some standard benchmarks. Along with this our model runs at $5$ FPS on Nvidia Quadro M1000M for VGA resolution images. Code is available at \href{https://github.com/aditya-vora/FCHD-Fully-Convolutional-Head-Detector}{https://github.com/aditya-vora/FCHD-Fully-Convolutional-Head-Detector}.
\end{abstract}
\section{Introduction and Related Work}
\label{sec:intro and related work}
Human head detection in crowded scenes is an important problem in the field of computer vision because of its wide range of applications. One such use-case could be head detection based crowd counting where detection based approach tends to give more reliable results compared to previous density map based crowd counting techniques \cite{zhang2015cross, zhang2016single, sam2017switching}. This is because, in case of density maps, it is not always the correct location which contributes to final crowd count. This leads to unreliable results especially in case of false positives. Several approaches for head detection has been proposed previously which are used in different contexts \cite{mae2005head, zhang2013multi, merad2010fast, vu2015context, stewart2016end, marin2014detecting}. \cite{merad2010fast} proposed human body segmentation approach in order to localize heads in a scene. \cite{subburaman2012counting} trained a sophisticated classifier based on the cascade of boosted integral features. In order to perform head detection based analytics, Marin et. al. \cite{marin2014detecting} firstly localized upper part of the human body and then trained a DPM \cite{felzenszwalb2010object} in order to detect head in the localized upper body part. The purpose of this head detection strategy was to find out how two characters are interacting in a movie scene. Recent years have witnessed various deep learning based head detectors \cite{vu2015context, stewart2016end}. One of the best performing head detection model \cite{vu2015context} extends RCNN detector for head detection using two types of contextual cues. First cue leverages the person-scene relations through a global CNN model. Second cue models the pairwise relations among objects using a structured output. Both of these cues are combined in a joint RCNN framework for head detection. Both the approaches for head detection i.e. \cite{marin2014detecting} and \cite{vu2015context} are designed in the context of movies where the average head size is large whereas average head count is small. These models when deployed for head detection in crowded scenes might fail to generalize because of the complete opposite scenario for head detection i.e. small average head size and large average head count. This is because no special care is taken to incorporate large scale and density variations in the previous models. This makes the entire task of head detection in crowded scenes a challenging one to be solved. Our work closely relates to \cite{stewart2016end} where they propose an anchor-free detection framework for head detection in crowded scenes which uses RNN in order to exploit information from deep representations in order to make predictions.

In this work, we propose an architecture for head detection especially for detection in crowded scenes. Our model is based on general anchor based detection pipeline like Faster-RCNN \cite{ren2017faster}, where we have a set of pre-set anchors which are generated by regularly tiling a collection of boxes with different scales in an image. However unlike a two-stage pipeline of Faster-RCNN (Region Proposal Network + ROI Pooling and Classification), in this architecture, we have only a single fully-convolutional network that can perform both classification and bounding box prediction. In this paper, we make the following contributions: 1) A fully convolutional, end-to-end trainable head detection model with systematic anchor selection strategy based on effective receptive field of the network which helps to achieve good results in crowded scenes. 2) Our model has very low memory requirements and inference time which makes it suitable for edge deployments. 3) We demonstrate that our model achieves comparable results to many other architectures as well as outperforming some of the previous approaches on challenging head detection datasets.

\section{Method}
\label{sec: method}
\subsection{Model architecture}
\begin{figure*}[!t]
\centering
\includegraphics[scale=0.25]{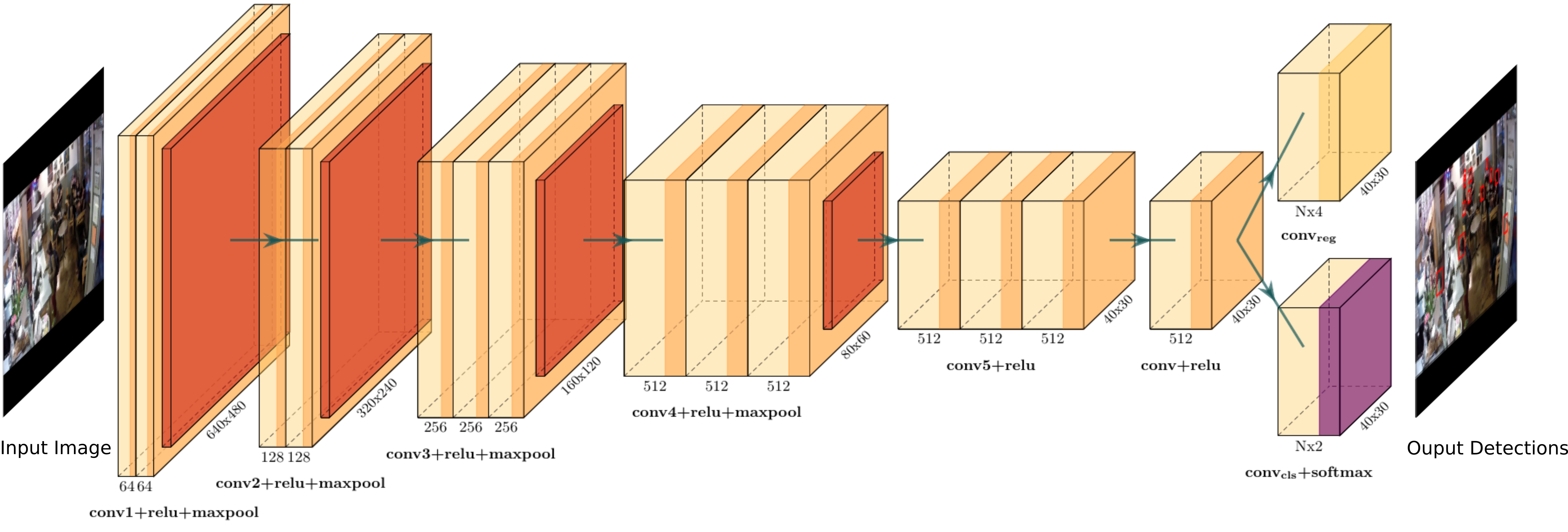}
\caption{Architecture of our head detection model. A set of pre-defined anchors are generated and a fully convolutional network is used to estimate the coordinates from the anchors alongwith the probability score. There are two heads to the model, 1) conv\textsubscript{reg}, $40\times30\times(N\times4)$. 2) conv\textsubscript{cls}, $40\times30\times(N\times2)$. Each $1\times1\times(N\times4)$ output from the conv\textsubscript{reg} will indicate the scale and shift of $N$ anchors at that location in the feature map. Here, $N=2$ which is the number of anchors.}
\label{fig: overall pipeline}
\end{figure*}

Deep architectures like \cite{krizhevsky2012imagenet, szegedy2015going, simonyan2014very} are capable of learning ``generalized features", which are useful in a variety of tasks. We start off by leveraging one of such already existing pre-trained VGG16 \cite{simonyan2014very}, as our base model for our architecture. We remove the final layers succeeding the conv5 layer of VGG16, and use the remaining weights as a starting point for our new training. The architecture is shown in Fig. \ref{fig: overall pipeline}. This pre-trained base model is appended with $3$ new convolutional layers: (1) The first conv layer (conv6) slides over the entire feature map in order to make detections at all locations. (2) The second conv layer (conv\textsubscript{reg}) is the regression head responsible for predicting localization coordinates. (3) The third conv layer (conv\textsubscript{cls}) is the classification head which predicts a probability score of a head. The conv\textsubscript{reg} and conv\textsubscript{cls} layers are implemented using a $1\times1$ convolutional kernel. The outputs from the conv\textsubscript{reg} undergo bounding box transformation, where the scale and shift predictions are converted to spatial coordinates. Number of conv\textsubscript{reg} and conv\textsubscript{cls} kernels depends on number of anchors at each pixel location $(N)$.

\subsection{Design of Anchor Scales}
Anchors are obtained by placing a pre-defined scale of boxes at regular intervals in an image. It is with respect to these anchors, the scales and shifts are predicted in an object recognition system \cite{ren2017faster}. We select the anchor scales based on the ``effective receptive field" rather than the ``theoretical receptive field". As pointed out by \cite{luo2016understanding}, a unit in a CNN has two types of receptive field, one is a theoretical receptive field and the other is the effective receptive field. Only a few center pixels of a theoretical receptive field is responsible for triggering the neurons in the final output. This center region of the theoretical receptive field is the effective receptive field. In the conv5 layer of VGG16, the theoretical receptive field size is $228$ pixels. Taking note of this, we prefer anchors of scales $2$ and $4$ which leads to anchor sizes $32\times32$ and $64\times64$ respectively. Anchor size is computed as $layer\_stride \times aspect\_ratio \times anchor\_scale$ where the $layer\_stride$ of the conv5 layer is $16$. Anchor sizes are approximately scaled down by $3.5$ times the theoretical receptive field which will match the effective receptive field size. Designing anchor scales in this manner helps to achieve good detection results especially in crowded scenes where the head size is small. This can be concluded from our experiments.

\subsection{Training}

\subsubsection{Approach}

Using our designed set of anchor scales, a set of predefined anchors are generated for each image after the initial preprocessing stage. In the preprocessing, we resize the image to $640\times480$ resolution and then do mean subtraction and normalization of the image. The anchors are tiled at a stride of $16$ because of $4$ max-pooling layers prior to conv5 layer. We select two anchors/pixel of size $32\times32$ and $64\times64$. Because of $4$ max-pooling layers prior to conv5 layer the resolution of the feature map obtained at the conv5 layer is of size $40\times30$ (as input resolution of the image is $640\times480$). Thus, for each image we obtain $2400$ anchors $(40\times30\times2)$. For each anchor, the network will predict $4$ regression coordinates through conv\textsubscript{reg} and $2$ probability scores through conv\textsubscript{cls}. Regression coordinates are in the form of scales and shifts the anchors need to undergo in order to localize an actual head in the scene. We do not consider the anchors going out of the boundary of the image for training. In order to assign labels to the anchors for training, we follow these $2$ strategies: (1) The anchor that have IoU $\geq0.7$ with ground-truth is labeled positive. (2) The anchors with maximum IoU with the ground-truth is assigned a positive label. We follow both strategies since in some cases the first condition might fail to assign a positive label to any of the anchors. Negative labels are assigned to those anchors which have IoU $\leq0.3$. The anchors that do not satisfy both the conditions are don't care and are not included in the training. We process a single image at a time during training. And from the anchors with assigned labels, we pick a total of $32$ positive and negative anchors in a ratio of $1:1$. 

\subsubsection{Loss Function}
The loss function used for the training of the model is a multi-task loss function, similar to that defined in training of an RPN by \cite{ren2017faster}. It is shown in Eq. \ref{eq:loss_function}.
\begin{equation}
    L(\{p_{i}\}, \{t_{i}\}) = \dfrac{1}{N_{cls}}\sum_{i}L_{cls}(p_{i},p_{i}^{*}) + \dfrac{1}{N_{reg}}\sum_{i}p_{i}^{*}L_{reg}(t_{i},t_{i}^{*})
\label{eq:loss_function}
\end{equation}
where $i$ is the index of chosen anchor which will range through $32$ selected anchors. $p_{i}$ is the predicted probability that an anchor $i$ contains head and $p_{i}^{*}$ is the ground-truth label i.e. $1$ or $0$. $t_{i}$ is the parameterized coordinates of the predicted bounding box (i.e. scale and shift), and $t_{i}^{*}$ is the parameterized coordinates of the ground truth. This parameterized coordinates are defined in the same manner as by \cite{girshick2015fast}. $L_{cls}$ denotes the classification loss, whereas $L_{reg}$ denotes the regression loss. $L_{cls}$ is computed over all the anchors whereas $L_{reg}$ is computed over only positive anchors and this is taken care by $p_{i}^{*}$ in front of $L_{reg}$ in the equation. $L_{cls}$ is the cross entropy loss. Whereas $L_{reg}$ is the smooth $L_{1}$ loss as defined by \cite{girshick2015fast}. Both the loss terms are normalized by $N_{cls}$ and $N_{reg}$ which are the number of samples accounted for classification and regression respectively.

\subsubsection{Hyperparameters}
The base model is initialized with pre-trained VGG16, which is trained using ImageNet dataset \cite{russakovsky2015imagenet}. All the layers of the pre-trained model along with new layers are retrained. The new layers are initialized with random weights sampled from a standard normal distribution with $\mu = 0$ and $\sigma = 0.01$. Weight decay is set to $0.0005$. The entire model is fine-tuned using SGD. Learning rate for training is set to $0.001$, and the model is trained for $15$ epochs ($\sim 160k$ iterations). We decay the learning rate with a scale of $0.1$ after completing $8$ epochs. Whole training was performed in PyTorch framework \cite{paszke2017automatic}.

\begin{table}[]
\begin{center}

\begin{tabular}{|c|c|c|c|}
\hline
\textbf{Dataset}           & \textbf{Scenario}                                                                                      & \textbf{Method}                                                              & \textbf{AP}   \\ \hline
\multirow{5}{*}{Brainwash} & \multirow{5}{*}{\begin{tabular}[c]{@{}c@{}}Crowded \\ scenes\\ (7.89 avg. \\ head count)\end{tabular}} & Overfeat - AlexNet                                                           & 0.62          \\ \cline{3-4} 
                           &                                                                                                        & ReInspect, $L_{fix}$                                                         & 0.60          \\ \cline{3-4} 
                           &                                                                                                        & ReInspect, $L_{firstk}$                                                      & 0.63          \\ \cline{3-4} 
                           &                                                                                                        & ReInspect, $L_{hungarian}$                                                   & 0.78          \\ \cline{3-4} 
                           &                                                                                                        & \textbf{Ours}                                                                & \textbf{0.70} \\ \hline
\multirow{5}{*}{Hollywood} & \multirow{5}{*}{\begin{tabular}[c]{@{}c@{}}Movie \\ scenes\\ (1.65 avg. \\ head count)\end{tabular}}   & DPM Face                                                                     & 0.37          \\ \cline{3-4} 
                           &                                                                                                        & R-CNN                                                                        & 0.67          \\ \cline{3-4} 
                           &                                                                                                        & Context - Local                                                              & 0.71          \\ \cline{3-4} 
                           &                                                                                                        & \begin{tabular}[c]{@{}c@{}}Context - Local+\\ Global + Pairwise\end{tabular} & 0.72          \\ \cline{3-4} 
                           &                                                                                                        & \textbf{Ours}                                                                & \textbf{0.74} \\ \hline
\end{tabular}
\end{center}
\caption{Evaluation of our model across various datasets.}
\label{tab: average_precision}
\end{table}

\section{Experiments}
\label{sec: experiments}
In order to have a thorough evaluation of our model, we benchmark our results on two publicly available datasets mentioned in Table \ref{tab: average_precision}. We train the model using the train set of the respective dataset (Brainwash-10461, Hollywood-216694) and then evaluate the model on the test set. Both the datasets are captured in different scenarios. Brainwash dataset\cite{stewart2016end} is captured in crowded scenes where the average head count is large ($7.89$) whereas HollywoodHeads dataset is collected from movie scenes where an average head count is small ($1.65$). Following settings are kept constant throughout the experiment. Anchor scales are set to $2$ and $4$. Non Maximum suppression (NMS) is performed during the test phase and not in the training phase. NMS is a kind of post-processing technique to filter out highly overlapping detections. In this technique, any two detections are suppressed to one if the IoU among them is greater than a threshold. The threshold for NMS is set to $0.3$. Following evaluation metrics is used, i.e. head is considered to be correctly detected if the IoU of the predicted bounding box with the ground truth is $\geq0.5$ \cite{everingham2015pascal}. In order to quantify the performance of our model, we plot the precision-recall curve along with computing average precision (AP). We then compare the performance of other baselines with our technique using this precision-recall curve and average precision table (Table \ref{tab: average_precision}).

\subsubsection{Results on Brainwash Dataset}
\begin{figure}
\centering
\includegraphics[scale=.34]{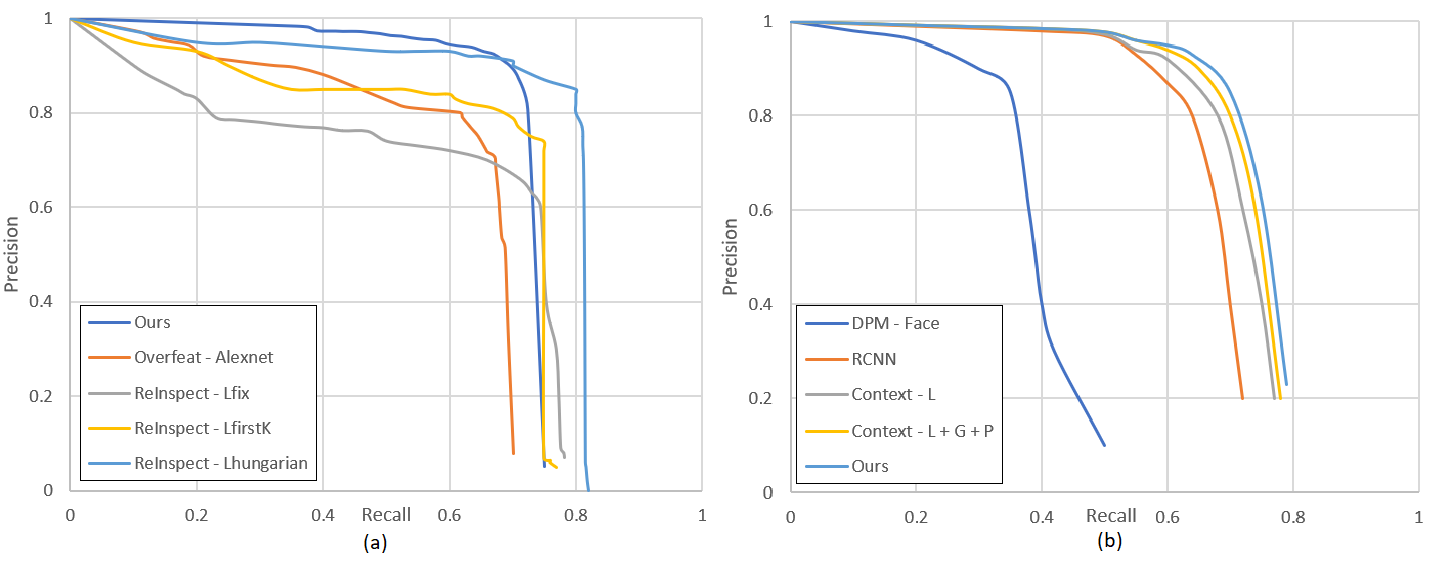}
\caption{Precision-Recall for Brainwash and HollywoodHeads dataset respectively.}
\label{fig: precision_recall}
\end{figure}

\begin{figure}
\centering
\includegraphics[scale=.5]{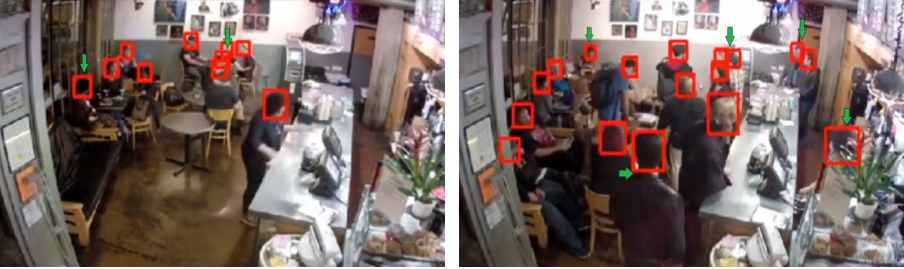}
\caption{Successful detection (green arrows) under challenging cases like high occlusion and high crowd density.}
\label{fig: success cases}
\end{figure}

\begin{figure}
\centering
\includegraphics[scale=.55]{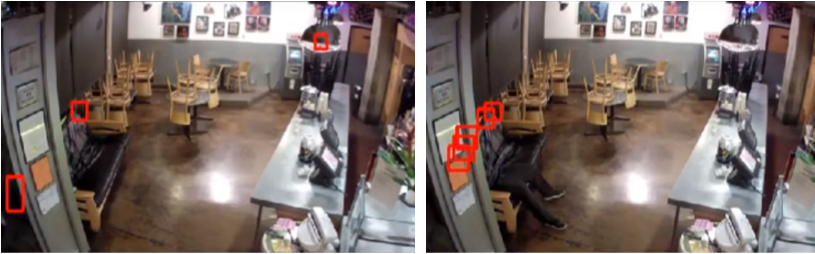}
\caption{Failure cases of our model. Gives false detections in scenarios where there is no or less heads to be detected.}
\label{fig: failure cases}
\end{figure}

We test our model using the test set of the Brainwash dataset which contains $484$ images. We consider same baselines for comparisons which were considered by \cite{stewart2016end} i.e. OverFeat with AlexNet architecture \cite{sermanet2013overfeat} and the current state-of-the-art model called ReInspect \cite{stewart2016end}. There are $3$ variants of ReInspect, which uses various types of losses i.e. $L_{fix}$, $L_{first}$, $L_{hungarian}$. Comparison with all of these variants is performed. Fig. \ref{fig: precision_recall}(a) shows the precision-recall curve on the test set. As can be seen from the figure, the precision-recall obtained by our technique is better than $3$ of the baselines i.e. Overfeat-AlexNet, ReInspect, $L_{fix}$ and ReInspect, $L_{firstk}$. ReInspect, $L_{hungarian}$ is the best performing model among all at EER (Equal Error Rate). Despite this, our model gives the best performance at a little lower recall i.e. best precision of $0.92$ at a recall of $0.65$ which can be observed from the precision-recall curve. AP attained by our model is $0.70$ which is better than $3$ baselines and comparable to best performing model (i.e. $0.78$). This can be seen in Table \ref{tab: average_precision}. Fig. \ref{fig: success cases} demonstrates some of the successful detection cases of our model under high occlusion and crowd density scenarios. Fig. \ref{fig: failure cases} shows some of the failure cases of our model.

\subsubsection{Results on HollywoodHeads Dataset}
In order to experiment with the HollywoodHeads dataset we train a completely fresh model using the train set. After training, the model is then tested on the test set which contains $1297$ images. We compare our approach with techniques used for comparison in \cite{vu2015context} i.e. DPM based head detector \cite{mathias2014face}, RCNN based head detector \cite{girshick2014rich}, and two variants of same technique i.e. 1) Local context model (Context-Local) and 2) Combined Local$+$Global$+$Pairwise context model (Context-L$+$G$+$P) as proposed in \cite{vu2015context}. Precision-recall curve is shown in Fig. \ref{fig: precision_recall}(b). Our model gives best performance compared to all the baselines. We achieve an AP of $0.74$ which is $\sim2\%$ higher compared to the state-of-the-art as shown in Table \ref{tab: average_precision}.

\begin{table}[]
\centering
\begin{tabular}{|c|c|c|}
\hline
\textbf{Platform}              & \textbf{Model} & \textbf{Speed} \\ \hline
\multirow{2}{*}{Quadro M1000M} & ReInspect      & 1fps           \\ \cline{2-3} 
                               & FCHD           & 5fps           \\ \hline
\multirow{2}{*}{Jetson TX2}    & ReInspect      & NA             \\ \cline{2-3} 
                               & FCHD           & 1.6fps           \\ \hline
\end{tabular}
\caption{Run-time comparison at VGA resolution}
\label{tab: timings}
\end{table}

\subsubsection{Timings}
In this section, we compare the time taken by our model for inference on different hardware platforms with previous approaches. Timings of our model are benchmarked on two platforms. The first platform is NVidia Quadro M1000M GPU which has 512 CUDA cores. The second platform is NVidia Jetson TX2 which has $256$ CUDA cores which is the most suitable device for edge deployments. We record the inference time of all the images in the test set and then the final timing is computed as the average of inference time of all the images in the test set. The timings are shown in Table. \ref{tab: timings}. As can be seen from the table that our model runs at 5 FPS which is $5\times$ faster than ReInspect on the same platform. Moreover, our model runs at 1.6 FPS on Jetson TX2 embedded platform where the other model even fails to load because of memory requirements. This is a major advantage of our model as along with being accurate, it is also suitable for edge deployments.

\begin{table}[]
\begin{center}
\begin{tabular}{|c|c|}
\hline
\textbf{Anchor Size}           & \textbf{AP} \\ \hline
$32\times32$, $64\times64$     & 0.70        \\ \hline
$64\times64$, $128\times128$   & 0.53        \\ \hline
$128\times128$, $256\times256$ & 0.45        \\ \hline
\end{tabular}
\end{center}
\caption{Effect of anchor size on AP}
\label{tab: anchor_scales}

\end{table}

\subsubsection{Ablation Experiments}
To make sure that the anchor size we choose for our experiments as well as for deployment are optimal, we do an ablation study where we train $3$ different models with varying anchor sizes i.e. (1) $32\times32$, $64\times64$ (default setting), (2) $64\times64$, $128\times128$, (3) $128\times128$, $256\times256$. These models are trained with the same hyperparameter settings. As it can be seen from Table \ref{tab: anchor_scales} that best AP is obtained when the anchor sizes are $32\times32$ and $64\times64$ which justifies our claim made in previous sections that, choosing anchor scales based on the effective receptive field gives better detection results.

\section{Conclusion}
\label{sec: conclusion}
In this work, we propose FCHD, an anchor based fully convolutional end-to-end trainable head detection model for crowded scenes. Our model achieves $0.70$ AP on challenging head detection dataset like Brainwash, which is comparable to previous techniques. Along with being accurate, it has very less run-time i.e. 5 FPS on Nvidia Quadro and 1.6 FPS on Jetson TX2. In the future, we plan to improve upon this work by getting better overall accuracy especially in cases where the head sizes are small. This could be achieved by combining the detections from various convolutional layers.

\bibliographystyle{IEEEbib}
\bibliography{strings,refs}

\end{document}